\documentclass[10pt,twocolumn,letterpaper]{article}

\usepackage{wacv}
\usepackage{times}
\usepackage{epsfig}
\usepackage{graphicx}
\usepackage{amsmath}
\usepackage{amssymb}
\usepackage{adjustbox}
\usepackage{multirow}
\usepackage{balance}

\def\wacvPaperID{362} 

\wacvfinalcopy 

\ifwacvfinal
\fi

\ifwacvfinal
\usepackage[breaklinks=true,bookmarks=false]{hyperref}
\else
\usepackage[pagebackref=true,breaklinks=true,colorlinks,bookmarks=false]{hyperref}
\fi

\begin{document}

\title{Time-Space Transformers for Video Panoptic Segmentation}

\author{Andra Petrovai\\
Technical University of Cluj-Napoca\\
Cluj-Napoca, Romania\\
{\tt\small andra.petrovai@cs.utcluj.ro}
\and
Sergiu Nedevschi\\
Technical University of Cluj-Napoca\\
Cluj-Napoca, Romania\\
{\tt\small sergiu.nedevschi@cs.utcluj.ro}
}

\maketitle

\ifwacvfinal
\thispagestyle{empty}
\fi

\begin{abstract}
 We propose a novel solution for the task of video panoptic segmentation, that simultaneously predicts pixel-level semantic and instance segmentation and generates clip-level instance tracks. Our network, named VPS-Transformer, with a hybrid architecture based on the state-of-the-art panoptic segmentation network Panoptic-DeepLab, combines a convolutional architecture for single-frame panoptic segmentation and a novel video module based on an instantiation of the pure Transformer block. The Transformer, equipped with attention mechanisms, models spatio-temporal relations between backbone output features of current and past frames for more accurate and consistent panoptic estimates. As the pure Transformer block introduces large computation overhead when processing high resolution images, we propose a few design changes for a more efficient compute. We study how to aggregate information more effectively over the space-time volume and we compare several variants of the Transformer block with different attention schemes. Extensive experiments on the Cityscapes-VPS dataset demonstrate that our best model improves the temporal consistency and video panoptic quality by a margin of 2.2\%, with little extra computation.
\end{abstract}

\section{Introduction}

Video panoptic segmentation \cite{kim2020video} extends panoptic segmentation \cite{panoptic_paper} to video and provides a holistic scene understanding across space and time by performing pixel level segmentation and instance level classification and tracking. The task has a broad applicability in many real-world systems such as robotics and automated driving, which naturally process video streams, rather than single frames. Video panoptic segmentation is more challenging than its image-level counterpart, because it requires temporally consistent inter-frame predictions.

\begin{figure}[t]
	\centering{\includegraphics[width=\columnwidth]{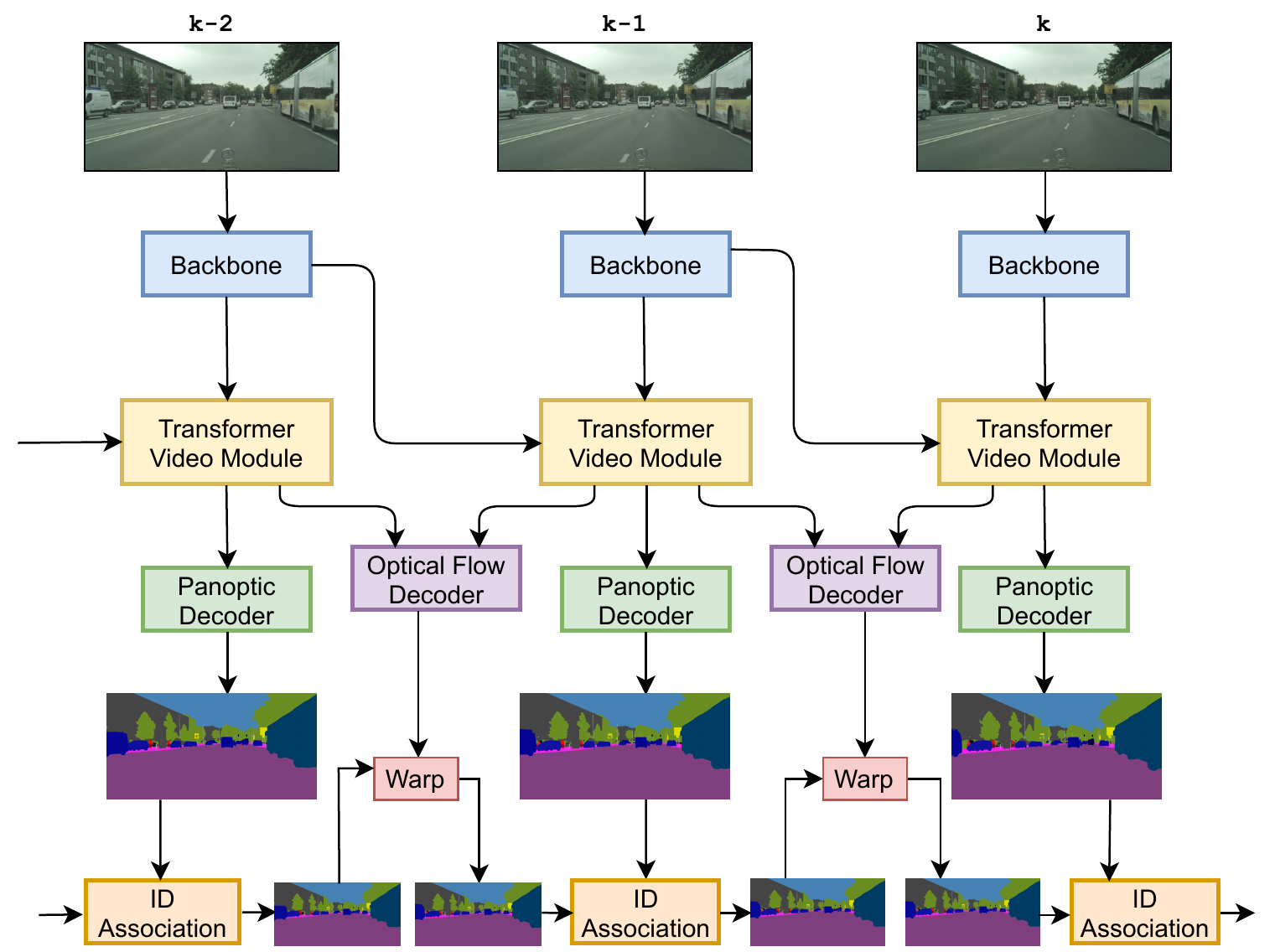}}
	\caption{High level overview of our VPS-Transformer network, which processes video frames and outputs panoptic segmentation and consistent instance identifiers. We propose a Transformer based video module to model temporal and spatial correlations among pixels from the current frame features and past frames features. Instance tracking is performed by warping the previous panoptic prediction with optical flow and associating the instance IDs with the current instance segmentation. }
\label{fig:highlevel}
\end{figure}

Video sequences provide rich information, such as temporal cues and motion patterns, which could be exploited for more accurate and consistent panoptic segmentation. Although we can clearly benefit from modeling the temporal correlations between frames, new challenges arise when processing video data. We can generally assume a high level of temporal consistency across consecutive frames, however, this could be broken by occlusions and new objects caused by fast scene evolution. In this context, temporal information should be carefully used such that we avoid introducing outdated information in our predictions \cite{nilsson2018semantic, jain2019accel}. On the other hand, extending a single frame solution to process multiple video frames often incurs a high computational cost \cite{kim2020video}, causing overhead in training and inference. However, striking a balance between efficiency and accuracy is important from a practical point of view. 

Unlike panoptic image segmentation \cite{panoptic_paper}, which has received increased attention from the research community \cite{PanopticFPN, PanopticDeepLab, DensePredictions}, video panoptic segmentation \cite{kim2020video} is a newly introduced task that has been less studied. Existing methods, such as VPSNet \cite{kim2020video} and ViP-DeepLab \cite{vip_deeplab} focus on improving the panoptic segmentation quality, however with a high computational cost. To increase consistency among frames, VPSNet  \cite{kim2020video} designs a temporal fusion module based on optical flow, that aggregates five neighboring past and future warped features. In contrast, our network is more accurate and efficient, although it operates in an online fashion: only the current frame is processed, while the Transformer block reads past features from the memory and attends to relevant positions in order to provide an enhanced representation. ViP-DeepLab \cite{vip_deeplab} models video panoptic segmentation as concatenated image panoptic segmentation and achieves the state-of-the-art for this task. Compared to this work, we explicitly encode spatio-temporal correlations for a boost in performance and obtain competitive results with a more lightweight network.

As shown in Figure \ref{fig:highlevel}, we propose a novel video panoptic segmentation approach by extending the single frame panoptic segmentation network Panoptic-DeepLab \cite{PanopticDeepLab} with a \textit{Transformer video module} and a \textit{motion estimation} decoder. Our module, inspired by the pure Transformer block \cite{vaswani2017attention}, refines the current backbone output features by processing the sequence of spatio-temporal features from current and past frames. As we strive to develop an efficient network, we design a \textit{lightweight variant of the pure Transformer block} that is faster than the original implementation \cite{vaswani2017attention}. We also present several ways to \textit{factorize the attention operation of the Transformer over space and time} and compare their accuracy and efficiency in extensive ablation studies. Given the enhanced feature representations from the Transformer module, three convolutional decoders recover the spatial resolution of the input image and multiple heads perform semantic segmentation, instance center prediction, instance offset regression and optical flow estimation. To ensure consistent instance identifiers for the same instance across frames, we implement a simple tracking module \cite{luiten2018premvos, weber2021step} based on mask propagation with optical flow and instance ID association between warped and predicted instance masks based on the class label and the intersection over union. Our video panoptic segmentation network is designed to achieve a good trade-off between speed and accuracy. Each newly introduced module is carefully designed to preserve the efficiency of the system. The proposed network can be trained in a weakly supervised regime with a sparsely annotated dataset, since it does not require labels for previous frames. We perform extensive experiments on the Cityscapes-VPS \cite{kim2020video} dataset and demonstrate that the proposed methods improve both image and video-based panoptic segmentation.

To summarize, our main contributions are the following:

\begin{enumerate}
\item{We propose a novel video panoptic segmentation network with a pure Transformer-based video module that applies spatial self-attention and temporal self-attention on sequences of current and past image features for more accurate panoptic prediction.}
\item{We extend our panoptic segmentation network with an optical flow decoder that is used for instance mask warping in the tracking process. Instance ID association between warped and predicted panoptic segmentation ensures temporally consistent instance identifiers across frames in a video sequence.}
\item{We propose a lightweight Transformer module inspired by the pure Transfomer architecture with three different designs of the attention mechanism: space self-attention, global time-space attention and local time-space attention. We compare the variants in extensive experiments.}
\item{We perform extensive experiments on the Cityscapes-VPS dataset and demonstrate that the proposed modules increase accuracy and temporal consistency, without introducing significant extra computational cost.}
\end{enumerate}


\section{Related Work}
\textbf{Panoptic Segmentation}  Proposal-based approaches employ an object detector for generating bounding box proposals, which are further segmented into instances. Semantic segments are usually merged with instance masks with a post-processing step that solves overlaps and semantic class conflicts. There are many works \cite{PanopticFPN,AndraITSC2019, li2018attention} that build their networks on top of the two-stage instance segmentation framework Mask-RCNN \cite{MaskRCNN}, which is extended with a semantic segmentation head. UPSNet \cite{UPSnet} proposes a parameter-free panoptic segmentation head supervised with an explicit panoptic segmentation loss. Seamless Segmentation \cite{SeamlessSegmentation} introduces a lightweight DeepLab-inspired segmentation head \cite{DeepLabV3+} and achieves high panoptic quality. Motivated by the recent success of one-shot detectors, FPSNet \cite{FPSNet} adopts RetinaNet \cite{lin2017focal} for proposal generation, and achieves fast inference speed. \cite{singleshot} extends RetinaNet with a semantic segmentation head and a pixel offset center prediction head, which is used for clustering pixels into instances. DenseBox \cite{DensePredictions} 
proposes a parameter-free mask construction method that reuses bounding box proposals discarded in the Non-Maxima Suppression step. Panoptic DeepLab \cite{PanopticDeepLab} achieves state-of-the-art results on multiple benchmarks with a bottom-up network that performs semantic segmentation and instance center regression. Axial-DeepLab \cite{wang2020axial} builds a stand-alone attention model with factorized self-attention along the height and width dimension.

\begin{figure*}[t]
	\centering{\includegraphics[width=\textwidth]{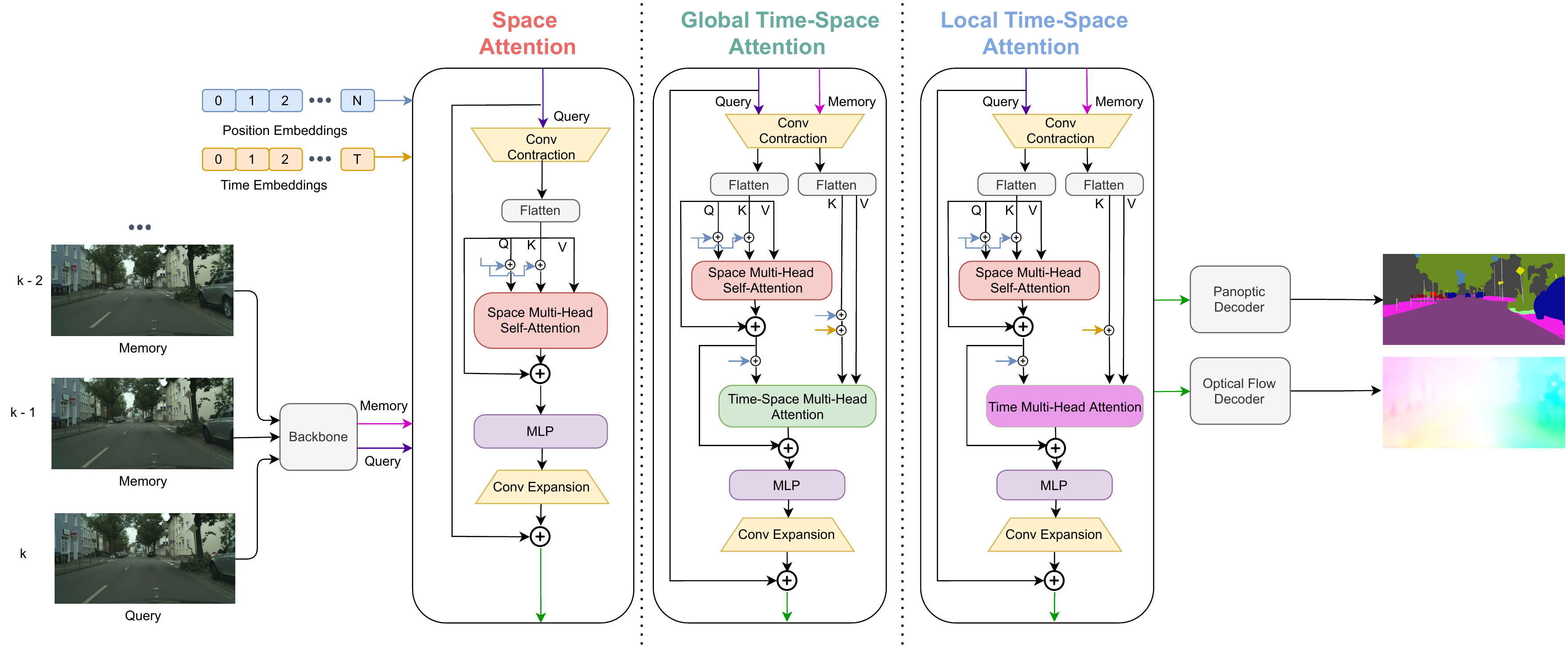}}
	\caption{We introduce a Transformer video module between the backbone and the decoders for more accurate prediction. We propose three variants of the module with various attention mechanisms: space attention, global time-space attention and local time-space attention. }
	\label{fig:lowlevel}
\end{figure*}

\textbf{Video Panoptic Segmentation} This task has been recently introduced in \cite{kim2020video}, which also proposes the baseline VPSNet network. VPSNet is built on top of the two-stage detector Mask R-CNN \cite{MaskRCNN} and the panoptic segmentation network UPSNet \cite{UPSnet}. The authors propose a module for pixel level fusion by gathering feature maps from the previous and next five frames, aligning them using optical flow and fusing them with spatio-temporal attention. Tracking is performed at object level with a MaskTrack head \cite{yang2019video}. ViP-DeepLab \cite{vip_deeplab} is a newly proposed network for video panoptic segmentation, which also performs monocular depth estimation. The authors extend the Panoptic-DeepLab \cite{PanopticDeepLab} network with a next-frame instance decoder which regresses the next-frame instance center offsets. A stitching algorithm based on mask IoU between region pairs is used for propagating instance identifiers from one frame to another. Concurrent work \cite{weber2021step} explores various tracking methods for panoptic video segmentation such as IoU association, SORT association \cite{bewley2016simple} and mask propagation with an external optical flow network \cite{teed2020raft}.

\textbf{Transformer} The Transformer network has been proposed in \cite{vaswani2017attention} for the task of machine translation. Recently, the Transformer architecture has been adopted for computer vision tasks. DETR \cite{carion2020end} employs the encoder-decoder architecture of the Transformer for object detection and demonstrates on par accuracy and run-time performance with the state-of-the-art CNN object detectors \cite{MaskRCNN}. ViT \cite{dosovitskiy2020image} applies Transformers on image patches and shows that when trained on large-scale datasets, it achieves state-of-the-art results for the task of image classification. Recently, the Transformer network has been adopted for the task of semantic segmentation \cite{zheng2020rethinking}. Self-attention with convolutional layers \cite{NonLocal, huang2019ccnet} has been explored in several works as a mechanism to increase accuracy for image-level recognition and temporal self-attention has been proposed for video object segmentation \cite{oh2019video} and semantic segmentation \cite{paul2021local}. In \cite{arnab2021vivit, bertasius2021space}, the authors explore several designs of a spatio-temporal Transformer network for video classification. Motivated by the recent success of the pure Transformer architecture in several tasks, especially video classification, we propose a novel Transformer-based video module in the context of a hybrid video panoptic segmentation network, where it can successfully model temporal dependencies between frames.

\section{Video Panoptic Segmentation Network}

We propose a video panoptic segmentation network, named VPS-Transformer, by extending Panoptic-DeepLab \cite{PanopticDeepLab} to perform panoptic segmentation and tracking of instances. In this section, we present the network architecture and implementation details.
\subsection{Model Architecture}

\begin{figure}[t]
	\centering{\includegraphics[width=\columnwidth]{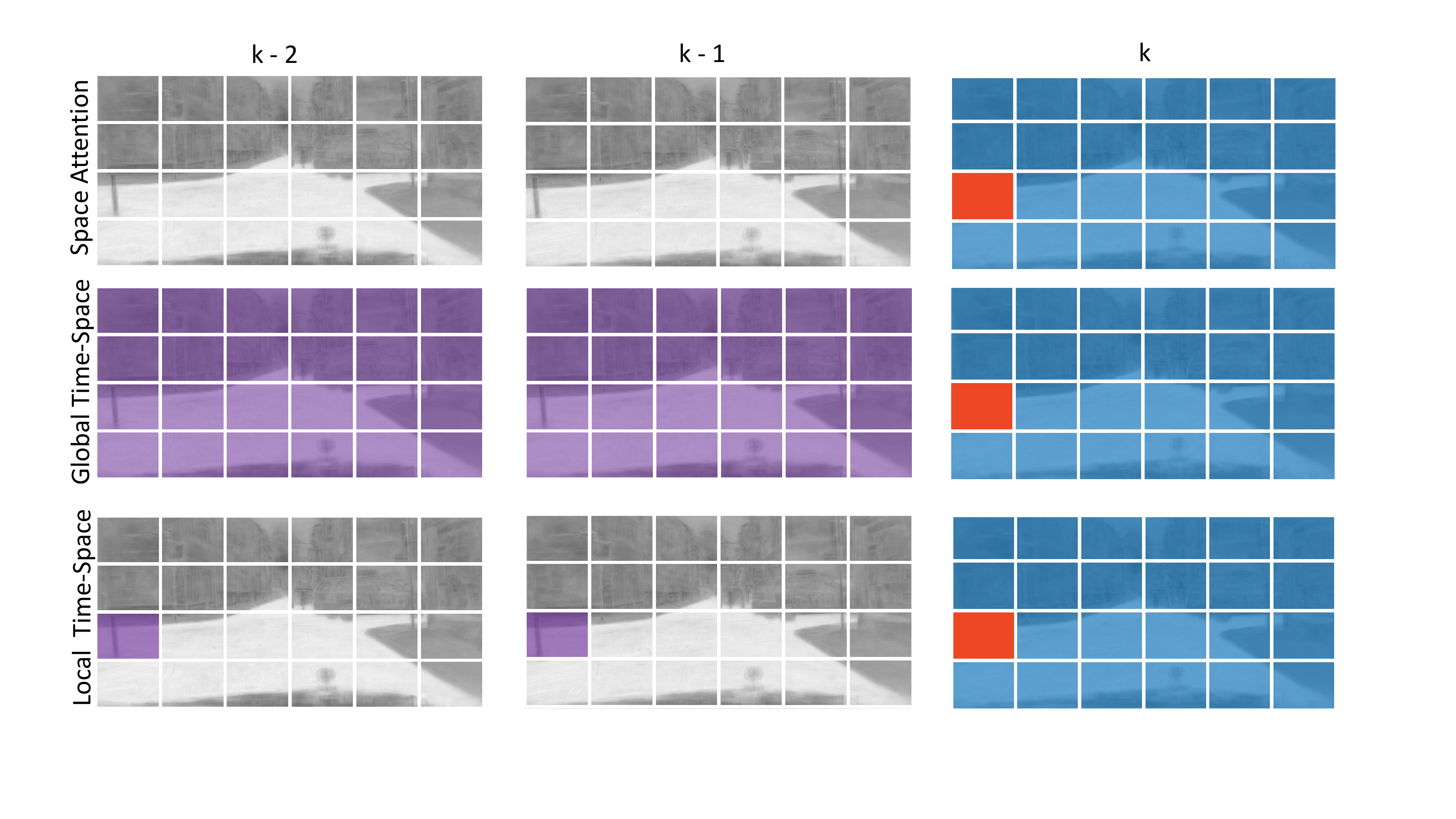}}
	\caption{The three attention schemes studied in this work. We present an example with three consecutive frames. The query token, shown in orange in frame $k$, attends to other tokens in the space and time dimension. When the space-time attention is factorized, we use blue for space and violet for time. We increase the granularity for visualization, but in practice each location from the sequence of features acts as a query.}
\label{features}
\end{figure}

\textbf{Baseline} We build our solution on top of the bottom-up image panoptic segmentation network Panoptic-DeepLab \cite{PanopticDeepLab}. The network regresses instance offsets and predicts semantic segmentation and instance centers. Class-agnostic instance segmentation is obtained by grouping foreground pixels to their closest center based on the predicted offsets. The final panoptic segmentation is generated by merging the semantic segmentation with the class-agnostic instance segmentation results by using a majority voting principle. The network consists of a shared backbone, dual decoders for semantic and instance segmentation and three heads for semantic, instance center and instance offset regression. We change the original implementation by reducing the output stride of the backbone from 16 to 32, and achieve similar panoptic quality and faster inference speed.

\textbf{Lightweight Transformer Video Module} Our network sequentially processes frames from a video sequence. A video module, plugged in between the backbone and the decoder, aggregates features of past frames in order to improve the current feature representation. We denote the past frames as \textit{memory} frames, while  the current frame is denoted as the \textit{query} frame. Our video module is inspired from the original Transformer architecture \cite{vaswani2017attention} with attention and self-attention mechanisms, as well as a multi-layer perceptron (MLP). 

The compute of the self-attention block scales quadratically with the size of the input as observed in \cite{tay2020efficient, srinivas2021bottleneck}, especially when processing high dimensional data. To reduce the complexity of the Transformer block, we wrap the Transformer between two pointwise convolutions in order to first reduce the number of channels and then to recover the original dimensionality, as seen in Figure \ref{fig:lowlevel}. In practice, we apply a $1 \times 1$ convolution to the backbone output features and map the number of channels from $C=2048$ to $d=1024$ and vice-versa. As the Transformer operates on sequences of input tokens of size $B \times N \times d$, we reshape the input query features \textbf{F} $\in  \mathbb R^{B \times H \times W \times d}$ to a sequence of flattened tokens \textbf{f} $\in  \mathbb R^{B \times HW \times d}$, where the first dimension is the batch dimension, the second one is the sequence length and the last one represents the embedding dimension. $B$ is the batch size, while $H$ and $W$ are the height and width of the feature maps. In the global time-space attention block, we map the memory features from the previous $T$ frames, \textbf{M} $\in  \mathbb R^{B \times T \times H \times W \times d}$ into a sequence of tokens \textbf{m} $\in  \mathbb R^{B \times THW \times d}$. In the local time-space attention design, we reshape the memory tokens to $ \mathbb R^{BT \times HW \times d}$. The spatial Transformer, which processes only the query frame, is further composed of a Multi-Head Self-Attention layer, Layer Normalization \cite{ba2016layer} and a 1-hidden-layer MLP.  When processing sequences of frames with the Time-Space Attention design, we add another temporal Multi-Head Attention layer that operates on the temporal dimension. After each attention module we employ residual connections and layer normalization \cite{ba2016layer}. We employ the same number of heads for all the Attention blocks. We obtain the best scores with one head configuration. The attention layer receives as input a query and a tuple of $(key, values)$ and returns a weighted sum of the values based on a similarity function of the query and key. In the self-attention module, the query, key and values come from the same source and are computed from the input \textbf{f} by linear projections with learnable weights $W^Q \in \mathbb R^{d \times d}$, $W^K \in \mathbb R^{d \times d}$, $W^V \in \mathbb R^{d \times d}$. Since the Transformer architecture lacks positional information, which is important for dense prediction, we inject learned position embeddings by adding them to the key and queries. In the time-space attention, we also inject learned temporal embeddings.

Given query, keys and values packed into \textbf{Q} $\in \mathbb R^{HW \times d}$, \textbf{K} $ \in \mathbb R^{HW \times d}$, \textbf{V}  $\in \mathbb R^{HW \times d}$, the attention operation can be formally defined as:

\begin{equation}
\text{Attention(\textbf{Q}, \textbf{K}, \textbf{V})} = \text{Softmax} \Big(\frac{\textbf{QK}^T}{\sqrt{d}}\Big)\textbf{V}
\end{equation}

A MLP with two fully connected layers and a GeLU \cite{hendrycks2016gaussian} activation in between gives the Transformer output sequence \textbf{o} $ \in  \mathbb R^{B \times HW \times d}$. The inner layer dimension of the MLP is equal to $d$. Finally, the output is reshaped to $\mathbb R^{B \times H \times W \times d}$ and the expansion pointwise convolution is applied. To obtain the refined feature representation, we apply a residual connection by adding the input, which is then followed by a ReLU non-linearity.

\textbf{Attention Variants}  We propose several attention schemes for the Transformer video module, as seen in Figure \ref{fig:lowlevel}. An example of how each scheme attends to different temporal and spatial tokens is given in Figure \ref{features}. Spatio-temporal attention in Transformer networks have also been recently explored for the video classification task \cite{arnab2021vivit, bertasius2021space}.

\textit{Space Attention} In this configuration the panoptic network only processes the query frame. We employ the Transformer module with spatial self-attention to model interactions between every position in the query frame features. This attention module performs $(H \cdot W)$ comparisons per query location.

\textit{Global Time-Space Attention} This module employs past frames features from the memory. The attention operation is factorized along the query space and memory time-space dimension. First, a spatial Multi-Head Self-Attention over the query is performed. Next, a Time-Space Multi-Head Attention module extracts global temporal correspondences between every position in the query and the memory. The query is given by the query frame while the key and values are given by the memory. For a query spatial position, $(H \cdot W\cdot T$) comparisons are needed in the Time-Space Multi-Head Attention.

\textit{Local Time-Space Attention} This module models spatio-temporal attention, but more efficiently than the previous module. After the spatial Multi-Head Self-Attention, we apply the Time Multi-Head Attention that computes attention locally on the temporal dimension only among the tokens located at the same spatial position. Time Multi-Head Attention performs  $(H \cdot W + T)$ comparisons for each query token.

\begin{table*}

\caption{Video panoptic segmentation results on the Cityscapes-VPS dataset with various Transformer Video Module variants. Each cell shows $ \text{VPQ}_k /\text{VPQ}_{k}^{Th} / \text{VPQ}_{k}^{St}$. VPQ is averaged over window size $k = \{1, 5, 10, 15\}$. $\text{VPQ}_1$ is equal to PQ. We vary the number of input frames to the Transformer Video Module from $S = 0$ to $S = 4$. With $S =0$ the network processes only the current frame.}
\label{all_variants}
\begin{adjustbox}{max width=\textwidth}
\begin{tabular}{lcccccc}
\hline 
\multicolumn{1}{l|}{\multirow{2}{*}{\textbf{Models}}}              & \multicolumn{4}{c|}{$ \text{VPQ}_k /\text{VPQ}_{k}^{Th} / \text{VPQ}_{k}^{St}$ for temporal window size $k$}                                                                                                                              & \multicolumn{1}{c|}{\multirow{2}{*}{VPQ}}        & \multirow{2}{*}{Time (ms)} \\ \cline{2-5}
\multicolumn{1}{l|}{}                                              & \multicolumn{1}{c|}{k = 1}              & \multicolumn{1}{c|}{k = 5}               & \multicolumn{1}{c|}{k = 10}             & \multicolumn{1}{c|}{k = 15}             & \multicolumn{1}{c|}{}                            &                            \\ \hline
\multicolumn{1}{l|}{Baseline (B) S = 0}                            & \multicolumn{1}{c|}{63.0 / 52.1 / 70.9} & \multicolumn{1}{c|}{51.1 / 27.3 / 68.5}  & \multicolumn{1}{c|}{48.0 / 21.2 / 67.5} & \multicolumn{1}{c|}{45.9 / 17.4 / 66.7} & \multicolumn{1}{c|}{\textbf{52.0 / 29.5 / 68.4}}          & 86                         \\
\multicolumn{1}{l|}{B + Tracking}                                  & \multicolumn{1}{c|}{63.0 / 52.1 / 70.9} & \multicolumn{1}{c|}{55.4 / 37.4 / 68.5}  & \multicolumn{1}{c|}{51.1 / 30.9 / 67.5} & \multicolumn{1}{c|}{49.9 / 27.0 / 66.7} & \multicolumn{1}{c|}{\textbf{55.1 / 36.8 /  68.4}}         & 100                        \\ \hline
                                                                   &                                         &                                          &                                         &                                         &                                                  &                            \\ \hline
\textbf{Local Time-Space Attention}                               & \multicolumn{1}{l}{}                    & \multicolumn{1}{l}{}                     & \multicolumn{1}{l}{}                    & \multicolumn{1}{l}{}                    & \multicolumn{1}{l}{}                             & \multicolumn{1}{l}{}       \\ \hline
\multicolumn{1}{l|}{B + Transformer Video Module  S = 1}           & \multicolumn{1}{c|}{64.8 / 54.9 / 72.0} & \multicolumn{1}{c|}{52.9 / 30.4 / 69.2}  & \multicolumn{1}{c|}{49.6 / 24.0 / 68.1} & \multicolumn{1}{c|}{47.5 / 20.4 / 67.1} & \multicolumn{1}{c|}{53.7 / 32.5 / 69.1}          & 97                         \\
\multicolumn{1}{l|}{B + Transformer Video Module S = 1 + Tracking} & \multicolumn{1}{c|}{64.8 / 54.9 / 72.0} & \multicolumn{1}{c|}{55.4 / 36.5 / 69.2}  & \multicolumn{1}{c|}{51.8 / 29.4 / 68.1} & \multicolumn{1}{c|}{49.8 / 26.0 / 67.1} & \multicolumn{1}{c|}{55.5 / 36.7 / 69.1}          & 111                        \\ \hline
\multicolumn{1}{l|}{B + Transformer Video Module  S = 2}           & \multicolumn{1}{c|}{64.7 / 54.7 / 72.0} & \multicolumn{1}{c|}{53.1 / 30.8 / 69.3}  & \multicolumn{1}{c|}{49.8 / 24.5 / 68.2} & \multicolumn{1}{c|}{47.7 / 20.8 / 67.2} & \multicolumn{1}{c|}{53.8 / 32.7 / 69.1}          & 98                         \\
\multicolumn{1}{l|}{B + Transformer Video Module S = 2 + Tracking} & \multicolumn{1}{l|}{64.7 / 54.7 / 72.0} & \multicolumn{1}{l|}{56.2 / 38.2 / 69.3}  & \multicolumn{1}{l|}{52.8 / 31.5 / 68.2} & \multicolumn{1}{l|}{50.6 / 27.8 / 67.2} & \multicolumn{1}{l|}{56.0 / 38.0 / 69.1}          & 112                        \\ \hline
\multicolumn{1}{l|}{B + Transformer Video Module S = 3}            & \multicolumn{1}{l|}{64.7 / 54.7 / 71.8} & \multicolumn{1}{l|}{53.0 / 30.9 / 69.1}  & \multicolumn{1}{l|}{49.7 / 24.4 / 68.0} & \multicolumn{1}{l|}{47.6 / 20.5 / 67.3} & \multicolumn{1}{l|}{\textbf{53.8 / 32.6 / 69.1}}          & 99                         \\
\multicolumn{1}{l|}{B + Transformer Video Module S = 3 + Tracking} & \multicolumn{1}{l|}{64.7 / 54.7 / 71.8} & \multicolumn{1}{l|}{57.4 / 41.1 / 69.1}  & \multicolumn{1}{l|}{54.2 / 35.0 / 68.0} & \multicolumn{1}{l|}{52.2 / 31.3 / 67.3} & \multicolumn{1}{l|}{\textbf{57.1 / 40.5 / 69.1}}          & 113                        \\ \hline
\multicolumn{1}{l|}{B + Transformer Video Module S = 4}            & \multicolumn{1}{l|}{64.6 / 54.6 / 71.9} & \multicolumn{1}{l|}{53.0 / 30.8 / 69.2}  & \multicolumn{1}{l|}{49.9 / 24.7 / 68.2} & \multicolumn{1}{l|}{47.7 / 21.0 / 67.0} & \multicolumn{1}{l|}{53.8 / 32.7 / 69.0}          & 100                        \\
\multicolumn{1}{l|}{B + Transformer Video Module S = 4 + Tracking} & \multicolumn{1}{c|}{64.6 / 54.6 / 71.9} & \multicolumn{1}{c|}{57.4 / 41.1 / 69.2}  & \multicolumn{1}{c|}{54.2 / 35.0 / 68.2} & \multicolumn{1}{c|}{52.0 / 31.2 / 67.0} & \multicolumn{1}{c|}{57.0 / 40.5 / 69.0}          & 114                        \\ \hline
                                                                   & \multicolumn{1}{l}{}                    & \multicolumn{1}{l}{}                     & \multicolumn{1}{l}{}                    & \multicolumn{1}{l}{}                    & \multicolumn{1}{l}{}                             & \multicolumn{1}{l}{}       \\ \hline
\multicolumn{7}{l}{\textbf{Global Time-Space Attention}}                                                                                                                                                                                                                                                                    \\ \hline
\multicolumn{1}{l|}{B + Transformer Video Module  S = 1}           & \multicolumn{1}{c|}{64.8 / 54.7 / 72.1} & \multicolumn{1}{c|}{53.3 / 31.0 / 69.4}  & \multicolumn{1}{c|}{49.9 / 24.6 / 68.3} & \multicolumn{1}{c|}{47.8 / 20.9 / 67.3} & \multicolumn{1}{c|}{\textbf{54.0 / 32.8 / 69.3}} & 98                         \\
\multicolumn{1}{l|}{B + Transformer Video Module S = 1 + Tracking} & \multicolumn{1}{c|}{64.8 / 54.7 / 72.1} & \multicolumn{1}{c|}{57.6 /  41.4 / 69.4} & \multicolumn{1}{c|}{54.4 / 35.2 / 68.3} & \multicolumn{1}{c|}{52.2 / 31.5 / 67.3} & \multicolumn{1}{c|}{\textbf{57.3 / 40.7 / 69.3}} & 112                        \\ \hline
\multicolumn{1}{l|}{B + Transformer Video Module  S = 2}           & \multicolumn{1}{c|}{64.7 / 54.8 / 72.0} & \multicolumn{1}{c|}{53.2 / 31.0 / 69.3}  & \multicolumn{1}{c|}{49.9 / 24.7 / 68.2} & \multicolumn{1}{c|}{48.0 / 21.2 / 67.3} & \multicolumn{1}{c|}{53.9 / 33.0 / 69.3}          & 101                        \\
\multicolumn{1}{l|}{B + Transformer Video Module S = 2 + Tracking} & \multicolumn{1}{l|}{64.7 / 54.8 / 72.0} & \multicolumn{1}{l|}{57.6 / 41.3 / 69.3}  & \multicolumn{1}{l|}{54.3 / 35.2 / 68.2} & \multicolumn{1}{l|}{52.2 / 31.3 / 67.3} & \multicolumn{1}{c|}{57.2 / 40.6 / 69.3}          & 115                        \\ \hline
\multicolumn{1}{l|}{B + Transformer Video Module S = 3}            & \multicolumn{1}{l|}{64.7 / 54.8 /  71.9}                   & \multicolumn{1}{l|}{53.1 / 30.9 / 69.3}                    & \multicolumn{1}{l|}{50.0 / 24.7 / 68.4}                   & \multicolumn{1}{l|}{47.9 / 21.2 / 67.4}                   & \multicolumn{1}{l|}{54.0 / 32.9 / 69.2}                            & 105                        \\
\multicolumn{1}{l|}{B + Transformer Video Module S = 3 + Tracking} & \multicolumn{1}{l|}{64.7 / 54.8 /  71.9}                   & \multicolumn{1}{l|}{56.0 / 37.7 / 69.3}                    & \multicolumn{1}{l|}{52.6 / 31.0 / 68.4}                   & \multicolumn{1}{l|}{50.5 / 27.2 / 67.4}                   & \multicolumn{1}{l|}{56.0 / 37.7 / 69.2}                       & 119                        \\ \hline
\multicolumn{1}{l}{}       \\ \hline
\multicolumn{7}{l}{\textbf{Space Attention}}                                                                                                                                                                                                                                                                                \\ \hline
\multicolumn{1}{l|}{B + Transformer Module  S = 0}           & \multicolumn{1}{c|}{64.5 / 54.3 / 71.9} & \multicolumn{1}{c|}{52.8 / 30.0 / 69.3}  & \multicolumn{1}{c|}{49.5 / 23.9 / 68.2} & \multicolumn{1}{c|}{47.4 / 20.2 / 67.2} & \multicolumn{1}{c|}{\textbf{53.6 / 32.1 / 69.2}} & 94                         \\
\multicolumn{1}{l|}{B + Transformer Module S = 0 + Tracking} & \multicolumn{1}{c|}{64.5 / 54.3 / 71.9} & \multicolumn{1}{c|}{57.2 /  40.5 / 69.3} & \multicolumn{1}{c|}{54.0 / 34.2 / 68.2} & \multicolumn{1}{c|}{51.7 / 30.3 / 67.3} & \multicolumn{1}{c|}{\textbf{56.8 / 39.8 / 69.2}} & 108                        \\ \hline
\end{tabular}
\end{adjustbox}
\end{table*}

\textbf{Instance Tracking.} We perform instance tracking by matching predicted instance masks with warped instance masks from the previous frame. Warping is performed using optical flow. We implement an optical flow decoder on top of the shared backbone. The first layer in the optical flow decoder is the correlation layer \cite{dosovitskiy2015flownet} computed between the previous frame features and the current frame features. The rest of the decoder follows a similar design to the instance decoder. For training the optical flow in an unsupervised setting, we employ the photometric loss that measures the photometric difference between the warped and the actual image. For ID association we follow the following algorithm. Given the panoptic segmentation at frame $t$ and the warped panoptic segmentation from frame $t - 1$ to frame $t$,  we compute the intersection over union (IoU) between every instance mask from the panoptic segmentation and the warped panoptic segmentation and vice-versa. For each instance, we store the corresponding instance with which it shares the semantic class and has the maximum IoU. Also, the IoU should be greater than a threshold, which we set to 0.3. Two pairs of instances are considered matched if they point to each other, that is, they have the maximum IoU. The instance identifier from the warped panoptic segmentation will be propagated to the matched instance in the current frame. Instances, that have not been matched, will receive a new instance identifier.

\subsection{Implementation Details}

We train and test our network on the Cityscapes-VPS dataset \cite{kim2020video}. Network training is done in three stages. In the first stage, the panoptic segmentation network is trained for image panoptic segmentation. We employ ImageNet pre-training \cite{deng2009imagenet}. For Panoptic-DeepLab, we train the network with a batch size of 8 and we generally follow the training settings from \cite{PanopticDeepLab}. Next, we freeze the backbone and the semantic and instance decoders, and train the optical flow decoder and the Transformer from scratch. The linear layers of the Transformer video are initialized by Xavier initialization \cite{glorot2010understanding}. In this stage, the network is trained using a minibatch of 8 images for a few epochs in order to get a rough initialization. We use the Adam optimizer, polynomial learning rate decay with a base learning rate of $1e-3$. We apply image augmentation, such as random horizontal flipping and random scaling with a factor in $[0.5, 2.0]$ from the original resolution $1024 \times 2048$. In the third stage, we fine-tune the CNN and the Transformer with fixed batch normalization layers. We set the learning rate to $1e-4$, and we train until convergence. Usually, our network converges within 30 epochs. 

\section{Experiments}

In this section, we provide experimental results on two datasets which provide pixel-level annotations, Cityscapes \cite{Cityscapes} and Cityscapes-VPS \cite{kim2020video}. We then discuss ablation studies and compare our results with the state-of-the-art for video panoptic segmentation.

\subsection{Experimental Setup} 

\textbf{Datasets} \textit{Cityscapes} \cite{Cityscapes} is an urban driving dataset with 5000 high-resolution images of size 1024 $\times$ 2048 pixels. The dataset is split into 2975 training, 500 validation and 1525 test images. Cityscapes provides instance-level annotations for 8 things classes and semantic-level annotations for 19 classes. 	

\textit{Cityscapes-VPS} \cite{kim2020video} extends Cityscapes \cite{Cityscapes} to panoptic video by providing annotations to every 5 frames from the 30-frame sequence in the validation set. The annotations in a video snippet are temporally aligned with consistent instance IDs across frames. The dataset provides 3000, 300 and 300, training, validation and test images.

\textbf{Evaluation metrics}  For panoptic segmentation, we adopt Panoptic Quality (PQ), Semantic Quality (SQ) and Recognition Quality (RQ) as defined in \cite{panoptic_paper}. Video panoptic segmentation is evaluated using the video panoptic quality (VPQ) metric, which measures image level quality, but also the temporal consistency across frames in a temporal window of size k=\{1, 5, 10, 15\}. While VPQ measures the  segmentation quality of both \textit{things} and \textit{stuff} classes, we also provide VPQ$^{Th}$ and  VPQ$^{St}$ scores, which are the VPQ averaged over the \textit{things} classes and over the \textit{stuff} classes, respectively.

\textbf{Inference time} We report the inference time of the network by measuring the forward pass of the network and the post-processing steps on a NVIDIA Tesla V100 GPU with a batch size of one. 

\subsection{Ablation Study}

\textbf{Baseline} The baseline network for our panoptic video segmentation network is the panoptic image segmentation network Panoptic-DeepLab \cite{PanopticDeepLab}. Since our video panoptic segmentation network includes extra modules such as the optical flow decoder and the Transformer video module, the inference time of the network increases. In order to preserve the efficiency of the video network, we first reduce the inference time of the image-level baseline network. We perform several experiments on the Cityscapes dataset using the baseline as seen in Table \ref{baseline}. The ResNet50 \cite{ResNet} backbone is used throughout this study. The original architecture of Panoptic-DeepLab \cite{PanopticDeepLab} features two ASPP modules with depthwise convolutions for context aggregation, dual decoders, for instance and semantic segmentation and a backbone output stride of 16. The network achieves a PQ score of 59.7\% with a reported time of the forward pass of 117 ms. By setting the output stride of the backbone to 32 and introducing another upsampling stage in the decoder we reduce the time to 86 ms and increase PQ with 1\%. In another experiment, we 
remove the instance segmentation decoder and place the instance center head and the offset regression head on top of the semantic segmentation decoder and get 59.8\% PQ. By replacing the depthwise convolutions in the ASPP decoder with simple convolutions, we increase PQ with 0.2\% and the inference time with 3 ms. The configuration that we use as the baseline is the one with dual decoders, depthwise convolutions in the ASPP and output stride of 32, which achieves the best trade-off between accuracy and efficiency.

\begin{table}[]
\caption{Ablation study of the baseline panoptic image segmentation Panoptic-DeepLab \cite{PanopticDeepLab} network on the Cityscapes \textit{val} set. We vary the number of decoders for instance and semantic segmentation, the type of  convolutions in the ASPP context module and the output stride of the backbone. }
\begin{adjustbox}{max width=\columnwidth}
\begin{tabular}{lcccc}
\hline
Decoder                 & OS & PQ   & mIoU  & Time (ms) \\ \hline
Dual + depthwise ASPP  \cite{PanopticDeepLab}            & 16 & 59.7 & -      & 117       \\
Dual + depthwise ASPP (our impl.)             & 16 & 60.5 & 79.6      & 111       \\
Single + depthwise ASPP & 32 & 59.8 & 79.0  & 82  \\
Single + ASPP           & 32 & 60.0 & 79.0  & 89    
     \\ \hline
Dual + depthwise ASPP   & 32 & \textbf{60.7} & \textbf{79.3} & \textbf{86}       
\\ \hline
\end{tabular}
\end{adjustbox}
\label{baseline}
\end{table}

\begin{figure*}[t]
	\centering{\includegraphics[width=\textwidth]{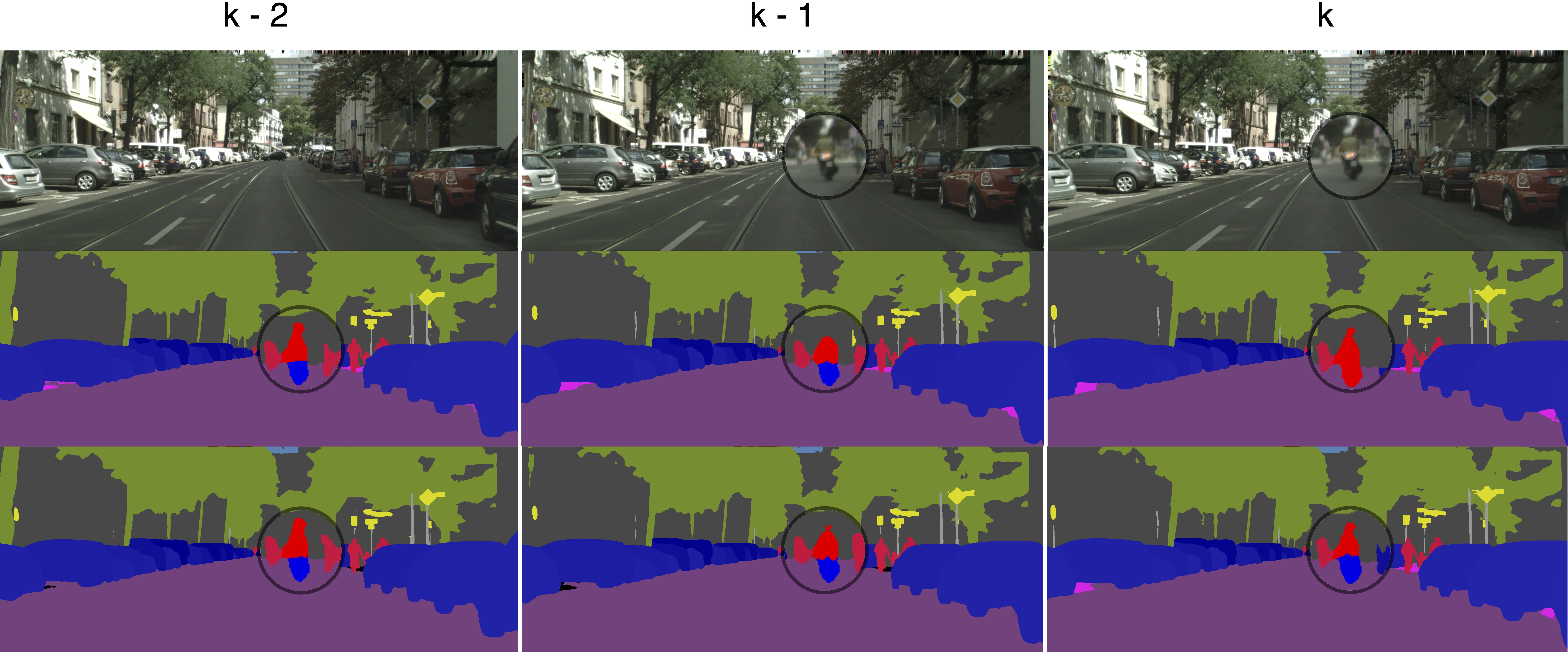}}
	\caption{Qualitative results on a sequence of three consecutive frames. On the second row we present the results of the baseline panoptic image segmentation nework, while on the third row we employ the Transformer video module with \textit{Global Time-Space Attention} with $S=1$. The encircled area with a rider and a motorcycle is zoomed in for better visualization. While the image-level panoptic segmentation network (second row) correctly segments the rider and motorcycle in the first two frames, it fails to segment the motorcycle in the third frame. We observe that the Transformer video module provides better temporal consistency and the rider and motorcycle are correctly segmented in all three frames (third row).}
\label{qualitativel}
\end{figure*}

\textbf{Attention Variants} We compare the proposed models with variants of the Transformer Video Module on the Cityscapes-VPS dataset in terms of accuracy and efficiency, in Table \ref{all_variants}. All models employ a ResNet50 backbone. The reported time is measured before and after tracking. For tracking, we measure the runtime of the network with an optical flow decoder, and also of the warping and the ID association algorithms. The baseline network, without the Transformer Video Module, performs single frame panoptic segmentation and achieves 63.0\% PQ and 52.0\% VPQ. Since video panoptic quality requires consistency between instance IDs across a video sequence, we observe a large accuracy drop when evaluating on a larger temporal window. Our tracking module solves instance IDs consistencies by propagating the IDs across frames and improves VPQ$^{Th}$ for all k greater than 1 with more than 3\%. The Space Attention model which introduces the Transformer Module with self-attention increases PQ by 1.5\%, VPQ by 1.6\% and VPQ after tracking by 1.7\%. In this variant, the Transformer Module proves to be a powerful mechanism to improve the current frame features only by spatial self-attention without using past information. The Space Attention model incurs a moderate computational cost of 12 ms over the baseline when processing high resolution images of 1024 $\times$ 2048 pixels. 
The Global Time-Space Attention models spatio-temporal correlations between every pair of pixels from the current frame and past frames features. This model yields the best scores among all the variants, and increases VPQ by 2\% and after tracking by 2.6\%, compared to the baseline. The results confirm that meaningful information from the past can be used to improve the current prediction. Also, the consistency between frames increases, thus enabling better ID association between instances in consecutive frames. A global view of the memory proves beneficial in this dense prediction task since it is not affected by the movement of instances between two frames. In terms of computational cost, this model requires slightly more compute than the Space Attention model since the Transformer module features an extra Multi-Head Attention block. The extra compute is also  determined by the increased number of input tokens, as the memory size increases. However, our best performing model variant uses only one past frame ($S=1$) and adds 14 ms to the baseline. The Local Time-Attention model is a slightly more efficient configuration of the Transformer Video Module. Since the Time Attention block computes attention temporally, among the tokens from the same spatial position, we observe a minimal change in inference speed when processing a higher number of frames. In order to achieve its best potential, this module requires a bigger memory of features. For $S=3$, the Local Time-Attention model increases the VPQ score with 1.8\% and 2\% after tracking compared to the baseline. For both Time-Attention modules, during inference when every frame is processed sequentially, we keep a memory of past backbone features in order to avoid feature re-computation.

\textbf{Memory Size} In this study, we analyze how the memory size affects the Global Time-Space Attention Transformer and the Local Time-Space Attention Transformer. We vary the memory size from 1 to 4 past frames, corresponding to $S=\{1, 2, 3, 4\}$ as seen in Table \ref{all_variants}. For both configurations, all memory sizes improve the VPQ compared to the baseline from 1.8\% to 2.0\% before tracking. The Local Time-Space Attention module benefits from long-term memory, achieving the best score for a memory of 3 frames. For this configuration, the tracking module further increases VPQ with 3.3\%, suggesting improved temporal consistency among frames. With a smaller memory of 1 or 2 frames, the tracking module has a slightly lower impact on the final score, with an improvement of 2.2\%. On the other hand, the Global Time-Space Attention memory can model complex correlations between pixels in the query frame and every pixel in the memory. This variant achieves the best scores when using a  short-term memory of 1 frame. For $S = 2$, the accuracy remains the same, however for bigger memories, the accuracy degrades, as the Transformer module is not capable of making very long-range global connections.  

\textbf{Transformer Configuration} We compare the model equipped with a pure Transformer \cite{vaswani2017attention} to our lightweight variant in Table \ref{transformer_variant}. The pure Transformer follows the original implementation, but uses no dropout. Our module is faster and provides increased video panoptic quality. 

In Table \ref{channels} we present the results of the Space Attention model with different number of channels in the Transformer module. In practice, we apply a $1 \times 1$ convolution on the backbone features in order to reduce the number of feature maps from 2048 to a lower dimension. All the other operations in the Transformer module, including the Attention and MLP blocks keep the dimensionality constant. We observe that projecting to a higher number of channels increases the accuracy.  Using $d=2048$ channels, we measure a time overhead of 21 ms. However, we select  $d = 1024$, since it achieves a good trade-off between accuracy and efficiency.

\begin{table}[]

\center
\caption{The proposed lightweight Transformer module versus the pure Transformer \cite{vaswani2017attention} with spatial self-attention only. We obtain higher video panoptic quality and inference speed.}
\label{transformer_variant}
\begin{adjustbox}{max width=\columnwidth}
\begin{tabular}{lcccccc}
\hline

Transformer module & k =1 & k = 5 & k = 10 & k = 15 & VPQ & Time (ms) \\ \hline
Pure Transformer \cite{vaswani2017attention} & 64.0 & 52.8 & 49.4 & 47.0 & 53.3 & 101 \\ 
Ours & \textbf{64.5} & \textbf{52.8} & \textbf{49.5} & \textbf{47.4} & \textbf{53.6} & \textbf{94} \\ \hline
\end{tabular}
\end{adjustbox}
\end{table}

\begin{table}[]

\center
\caption{The effect of varying the number of channels in the Transformer Video Module with \textit{Space Attention}. We report the total inference time of the model and the time overhead over the baseline on the Cityscapes-VPS \textit{val} set without tracking. }
\label{channels}
\begin{adjustbox}{max width=\columnwidth}
\begin{tabular}{lccccc}
\hline
Channels & mIoU & PQ & VPQ & $\Delta$T (ms) & Time (ms) \\ \hline
 512 & 78.9 & 63.8 & 52.7  & +4 & 90  \\
 768 & 79.1 & 64.0 & 52.9 & +6 & 92 \\
1024 & \textbf{79.8}  & \textbf{64.5} & \textbf{53.6} & +8 & 94  \\ \hline
\end{tabular}
\end{adjustbox}
\end{table}

\textbf{Qualitative Results} In Figure \ref{qualitativel} we provide a comparison between the baseline panoptic image segmentation network and the proposed video counterpart. In this example, the baseline panoptic segmentation network is not able to correctly segment the motorcycle in the last frame. However, by using the Transformer video module the temporal consistency is increased and the motorcycle is correctly segmented over time.

\subsection{Comparison to the State-of-the-Art}

 In Table \ref{stateoftheart}, we compare our results with the state-of-the-art on the Cityscapes-VPS dataset \cite{kim2020video}. VPSNet \cite{kim2020video} and our VPS-Transformer network are pretrained on the Cityscapes \textit{fine} train dataset \cite{Cityscapes}. Although VPSNet fuses features from 10 neighboring frames, from both past and future, we achieve better performance  with 1.2\% higher VPQ using our model with ResNet50 backbone \cite{ResNet}. At the same time, our network is 7$\times$ faster. With a more powerful backbone, HRNet-W48 \cite{WangSCJDZLMTWLX19}, we surpass VPSNet with 3.7\% VPQ, while still being 4 $\times$ faster. ViP-DeepLab \cite{vip_deeplab} is a network that extends Panoptic-DeepLab \cite{PanopticDeepLab} and solves the task of depth-aware video panoptic segmentation. We do not have access to the results of the ViP-DeepLab network pretrained on the Cityscapes fine train dataset for a fair comparison. Therefore, we present the ViP-DeepLab results of the network pretrained on a much larger set of images, on Mapillary Vistas dataset \cite{Mappilary} and Cityscapes \cite{Cityscapes} train dataset with pseudo-labels, as this is the most similar to our setting. ViP-DeepLab does not disclose the inference time, but we approximate it to be more than 400 ms, which is the time of its baseline panoptic image segmentation network Panoptic-DeepLab with WR-41 \cite{chen2020naive}. ViP-DeepLab extends its baseline by adding an ASPP module \cite{DeepLabV3+}, a next-frame instance decoder and a next-frame instance center regression head. Although our VPS-Transformer network with a HRNet-W48 backbone does not enjoy the same powerful pretraining, we achieve similar VPQ scores, 59.8\% vs 59.9\% VPQ. However, our VPS-Transformer network runs at least twice faster than ViP-Deeplab with an inference time of 185 ms. Our network shows a good trade-off between speed and accuracy and the potential to be used in real-world applications, such as automated driving.

\begin{table}[]

\caption{Comparison to state-of-the-art video panoptic segmentation networks on the Cityscapes-VPS \textit{val} set. VPSNet and our VPS-Transformer are pretrained on the Cityscapes \textit{fine} train set. ViP-DeepLab is pretrained on Mapillary Vistas (MV) and on Cityscapes \textit{fine} train set extended with pseudo-labels. ViP-DeepLab* results are obtained by running the model provided by the authors in \cite{vip}.}
\label{stateoftheart}
\begin{adjustbox}{max width=\columnwidth}
\begin{tabular}{llccc}
\hline
Method        & Backbone  & PQ       &  VPQ & Time (ms)            \\ \hline
VPSNet \cite{kim2020video}   & ResNet50  & 62.7  & 56.1  & 770                   \\ 
ViP-DeepLab* \cite{vip}    & ResNet-50  & 60.6 & 52.8  & -    \\
ViP-DeepLab + MV \cite{vip_deeplab}    & WR-41  & 67.9 & 59.9  & $>$ 400                    \\ 
\hline
Baseline - Panoptic DeepLab \cite{PanopticDeepLab}      & ResNet50            &    63.0   & 52.0    & 86                   \\
VPS-Transformer  (ours)     & ResNet50                     & 64.8  & 57.3  & 112    \\ \hline
Baseline - Panoptic DeepLab \cite{PanopticDeepLab}      & HRNet-W48           &    66.1   & 55.1    & 168                   \\
VPS-Transformer (ours)  & HRNet-W48  & 67.6  & 59.8  & 185    \\ \hline
\end{tabular}
\end{adjustbox}
\end{table}

\section{Conclusions}
In this paper, we introduce a novel video module inspired by a pure Transformer block to model spatio-temporal correlation between consecutive frames in the context of video panoptic segmentation. The proposed module can aggregate past features from the memory in order to refine the current panoptic prediction and increase temporal consistency. We introduce a lightweight video module design with several attention mechanism variations and perform extensive experiments and comparisons in terms of accuracy and efficiency. Finally, we implement an instance tracking module with optical flow and instance ID association. Experiments demonstrate that the our network achieves considerably better video panoptic quality and temporal consistency without introducing a significant extra computational cost. 

\section*{Acknowledgment}
This work was supported by the SEPCA project PN III PCCF no. 9/2018.

{\small
\bibliographystyle{ieee_fullname}
\bibliography{egbib}

}

\end{document}